\documentclass[runningheads]{llncs}
\usepackage{graphicx}
\usepackage{amsmath}
\usepackage{mathtools}
\usepackage{adjustbox}
\usepackage{parskip}
\usepackage{amssymb}
\DeclareMathOperator*{\argmax}{arg\,max}

\usepackage{changepage}
\usepackage{comment}
\usepackage{enumerate}
\usepackage[font=small,labelfont=bf]{caption}

\usepackage[ruled,vlined,algo2e]{algorithm2e}

\usepackage{comment}
\SetKwInput{KwInput}{Input}  
\usepackage{algorithm}
\usepackage{algorithmic}

\usepackage{array}
\newcolumntype{L}{>{$}l<{$}}
\newcolumntype{C}{>{$}c<{$}}
\newcolumntype{R}{>{$}r<{$}}

\usepackage{booktabs}       % professional-quality tables
\usepackage{subcaption}

\usepackage{xcolor}
\begin{document}
\title{Transferable Deep Metric Learning for Clustering}

\author{Simo Alami.C \inst{1,2} \and
Rim Kaddah\inst{2}\and
Jesse Read\inst{1}}
\authorrunning{S.Alami et al.}

\institute{Department of Computer Science, Ecole Polytechnique, Palaiseau, France
\email{\{mohamed.alami-chehboune, jesse.read\}@polytechnique.edu}\\
\and
IRT SystemX, Palaiseau, France\\
\email{rim.kaddah@irt-systemx.fr}}
\maketitle              % typeset the header of the contribution
\begin{abstract}
Clustering in high dimension spaces is a difficult task; the usual distance metrics may no longer be appropriate under the curse of dimensionality. Indeed, the choice of the metric is crucial, and it is highly dependent on the dataset characteristics. However a single metric could be used to correctly perform clustering on multiple datasets of different domains. We propose to do so, providing a framework for learning a transferable metric. We show that we can learn a metric on a labelled dataset, then apply it to cluster a different dataset, using an embedding space that characterises a desired clustering in the generic sense. We learn and test such metrics on several datasets of variable complexity (synthetic, MNIST, SVHN, omniglot) and achieve results competitive with the state-of-the-art while using only a small number of labelled training datasets and shallow networks.

\end{abstract}

\section{Introduction}\label{intro}
%context
Clustering is the unsupervised task of assigning a categorical value $y_i \in \{1,\ldots,k\}$ to each data point $x_i \in \mathbf{X}$, where no such example categories are given in the training data; i.e., we should map
$\mathbf X= \{x_1,\ldots,x_n\}\mapsto \mathbf Y = \{y_1,\ldots,y_n\}$
	%\underbrace{\{x_1,\ldots,x_n\}}_{\bf X \in \mathbb{R}^{n \times d}} \mapsto \underbrace{\{y_1,\ldots,y_n\}}_{\bf y \in \mathbb{N}^n}
with $\bf X$ the input matrix of n data points, each of dimension d; where $y_i = \kappa$ implies that data point $x_i$ is assigned to the $\kappa$-th cluster. 

Clustering methods complete this task by measuring similarity (the distance) between training pairs, using a similarity function  $s(x_i,x_j) \in \mathbb{R}_+$. 
This similarity function should typically reflect subjective criteria fixed by the user. Basically, this means that the user decides what makes a good clustering. As mentioned in \cite{learn}, ``since classes are a high-level abstraction, discovering them automatically is challenging, and perhaps impossible since there are many criteria that could be used to cluster data (e.g., we may equally well cluster objects by colour, size, or shape). Knowledge about some classes is not only a realistic assumption, but also indispensable to narrow down the meaning of clustering". Taking the example of MNIST \cite{MNIST_digits}, one usually groups the same numbers together because these numbers share the highest amount of features (e.g., mutual information based models do that). However one may want to group numbers given their roundness. In this case, we may obtain two clusters, namely straight shaped numbers (i.e., 1, 4,7) and round shaped numbers (i.e., all the others). Both clustering solutions are relevant, since each clustering addresses a different yet possible user subjective criteria (i.e., clustering semantics).

Finding an automated way to derive and incorporate user criteria in a clustering task based on intended semantics can be very hard. Nowadays, the wide availability of shared annotated datasets is a valuable asset and provides examples of possible user criteria. Hence, we argue that, given ``similar'' annotated data, classification logic can be used to derive a user criteria that one can apply to clustering similar non-annotated data. For example, we consider the situation where a human is placed in front of two datasets, each one consisting of letters of a certain alphabet she does not understand. The first dataset is annotated, grouping the same letters together. Only by seeing the first dataset, the person can understand the grouping logic used (grouping same geometrical shapes together) and replicate that logic to the second non annotated dataset and cluster correctly its letters. 

%problem statement
In this paper, we are interested in tackling the problem of clustering data when the logic (i.e., user clustering criteria) is encoded into some available labelled datasets.  
This raises two main challenges, namely (1) find a solution that works well on the classification task but (2) ensure transferability in its decision mechanism so it is applicable to clustering data from a different domain. 

%reasoning steps to building the solution
We believe that addressing these challenges calls for the design of a scoring function that should be as general as possible to ensure transferability but is specific enough not to miss the user criteria. More specifically, the scoring function should be a comparing the logic used to produce a certain clustering to the one used to produce clusterings of the already seen training datasets. Using the concept of logic is useful as a logic is general enough to be used on any dataset and specific enough as is it is the main common property shared by all training dataset. Our goal is then to find a suitable metric that retrieves and encapsulate the seen concept for scoring a clustering outcome.

Moreover, modern applications require solutions that are effective when data is of high dimension (i.e., large $d$). While distance-based approaches are broadly used for clustering (e.g., Euclidean distance), we argue that they are not suitable for our problem since they would yield in data specific models in addition to their poor performance in high dimensional spaces due to the curse of dimensionality. 
To lower dimensionality, a solution is to perform instance-wise embeddings $x_i \mapsto z_i$, e.g., with an autoencoder. However this mechanism is still domain specific. 

To achieve training on more general patterns, we think it is necessary to take the dataset in its entirety. Therefore, instead of learning a metric that compares pairs of data points in a dataset instance (like a similarity measure), a learned metric is applied to sets of data points so comparison is done between sets. The metric can be intuitively understood as a distance between the logic underlying a given clustering and the general logic that was used to produce clusterings in training datasets.

%explain the need of embedding -> for transferability, to standardize data format and to tackle high dimensionaity
For this, we propose a solution where we use a graph autoencoder \cite{GAE} to embed a set of data points into a vector of chosen dimension. Then, we use the critic part of a Wasserstein GAN (WGAN) \cite{WGAN} to produce a continuous score of the embedded clustering outcome. This critic represents the metric we seek. Thus, our main contributions are:
\vspace{-2mm}
\begin{itemize}
    \item We provide a framework for joint metric learning and clustering tasks.
    \vspace{-2mm}
    \item We show that our proposed solution yields a learned metric that is transferable to datasets of different sizes and dimensions, and across different domains (either vision or tabular) and tasks. 
    \vspace{-2mm}
    \item We obtain results competitive to the state-of-the-art with only a small number of training datasets, relatively simple networks, and no prior knowledge (only an upper bound of the cluster number that can be set to a high value).
    \vspace{-6mm}
    \item Our method is scalable to large datasets both in terms of number of points or dimensions (e.g the SVHN dataset used in section \ref{sec:experiments}) as it does not have to compute pairwise distances and therefore does not heavily suffer when the number of points or dimensions increase. 
    \vspace{-2mm}
    \item We test the metric on datasets of varying complexity and perform on par with the state-of-the-art while maintaining all the advantages cited above.
\end{itemize}

\section{Related Work}\label{related}  

Using auto-encoders before applying classic clustering algorithms resulted in a significant increase of clustering performance, while still being limited by these algorithms capacity.
%DEC gets rid of on-top algorithms but still has limitations
Deep Embedding Clustering (DEC) \cite{DEC} gets rid of this limitation at the cost of more complex objective functions. It uses an auto-encoder along with a cluster assignment loss as a regularisation. The obtained clusters are refined by minimising the KL-divergence between the distribution of soft labels and an auxiliary target distribution. DEC became a baseline for deep clustering algorithms. Most deep clustering algorithms are based on classical center-based, divergence-based or hierarchical clustering formulations and hence bear limitations like the need for an \textit{a priori} number of clusters.

% Other stuff exists too but it's not transferable and therefore of no interest to us
MPCKMeans \cite{mpckmeans} is more related to metric learning as they use constraints for both metric learning and the clustering objective. However, their learned metrics remain dataset specific and are not transferable.

% Except \cite{transfer_clutsering} -- which we use as a baseline.

Constrained Clustering Network (CCN) \cite{transfer_clustering}, learns a metric that is transferable across domains and tasks. Categorical information is reduced to pairwise constraints using a similarity network. Along with the learned similarity function, the authors designed a loss function to regularise the clustering classification. But, using similarity networks only captures local properties instance-wise rather than global geometric properties of dataset clustering. Hence, the learned metric remains non fully transferable, and requires to adapt the loss to the domain to which the metric is transferred to. 

In Deep Transfer Clustering (DTC) \cite{learn} and Autonovel \cite{autonovel}, the authors tackle the problem of discovering novel classes in an image collection given labelled examples of other classes. They extended DEC to a transfer learning setting while estimating the number of classes in the unlabelled data. Autonovel uses self-supervised learning to train the representation from scratch on the union of labelled and unlabelled datasets then trains the data representation by optimizing a joint objective function on the labelled and unlabelled subsets of data. We consider these two approaches as our state of the art baselines. 

\section{Our Framework}

To restate our objective, we seek an evaluation metric 
\begin{equation}\label{eq:map2}
    \begin{split}
        r : \mathbb R^{\bf n\times d} \times \mathbb {N}^{\bf n}\rightarrow \mathbb{R}\\
        (\bf X,\bf y)\mapsto r(\bf X,\bf y)
    \end{split}
\end{equation}
where $\bf X \in \mathbb R^{n\times d}$ is a dataset of $n$ points in $d$ dimensions and $\bf y \in \mathbb N^n$ a partition of $\bf X$ (i.e. a clustering of $\bf X$). Metric $r$ should provide a score for \emph{any} labelled dataset of any dimensionality; and in particular this score should be such that $r(\bf{X},\bf y)$ is high when the hamming distance between the ground truth labels $\bf y^*$ and $\bf y$ is small (taking cluster label permutations into account). This would mean that we could perform clustering on any given dataset, simply by solving an optimisation problem even if such a dataset had not been seen before. 

Formally stated, our goal is: (1) to produce a metric $r$ that grades the quality of a clustering such that $\bf{y}^*=\argmax_{\bf y} r(\bf X, \bf y)$; (2) Implement an optimisation algorithm that finds $\bf y^*$; (3) use (1) and (2) to perform a clustering on a new unrelated and unlabelled dataset. 
%For the sake of simplicity the function $r$ will be noted $r(\mathcal{X}_i,\mathcal{Y}_i)$ or $r(z)$, as $z_{(\mathcal{X}_i,\mathcal{Y}_i)}$ is an encoding of $(\mathcal{X}_i,\mathcal{Y}_i)$. 
We use a collection $\mathcal{D} = \{\mathbf{X}_l,\mathbf{y}_l^*\}_{l=1}^\ell$ of labelled datasets as examples of correctly `clustered' datasets, and learn $r$ such that $\mathbb{E}[r(\mathbf{X},\mathbf{y})]$ is high. In order to make $r$ transferable between datasets, we embed each dataset with its corresponding clustering ($\mathbf X_l,\mathbf y_l)$ into a vector $\mathbf z_l \in \mathbb R^{\bf e}$. More formally, the embedding function is of the form:
\begin{equation}
\begin{split}
    g: \, \, & \mathbb R^{\bf n\times d}\times \mathbf Y \rightarrow \mathbb R^\mathbf e \\
    & (\bf X,\bf y)\mapsto \bf z
\end{split}
\end{equation}

\begin{figure}[h!]
    \centering
    \includegraphics[width=\columnwidth]{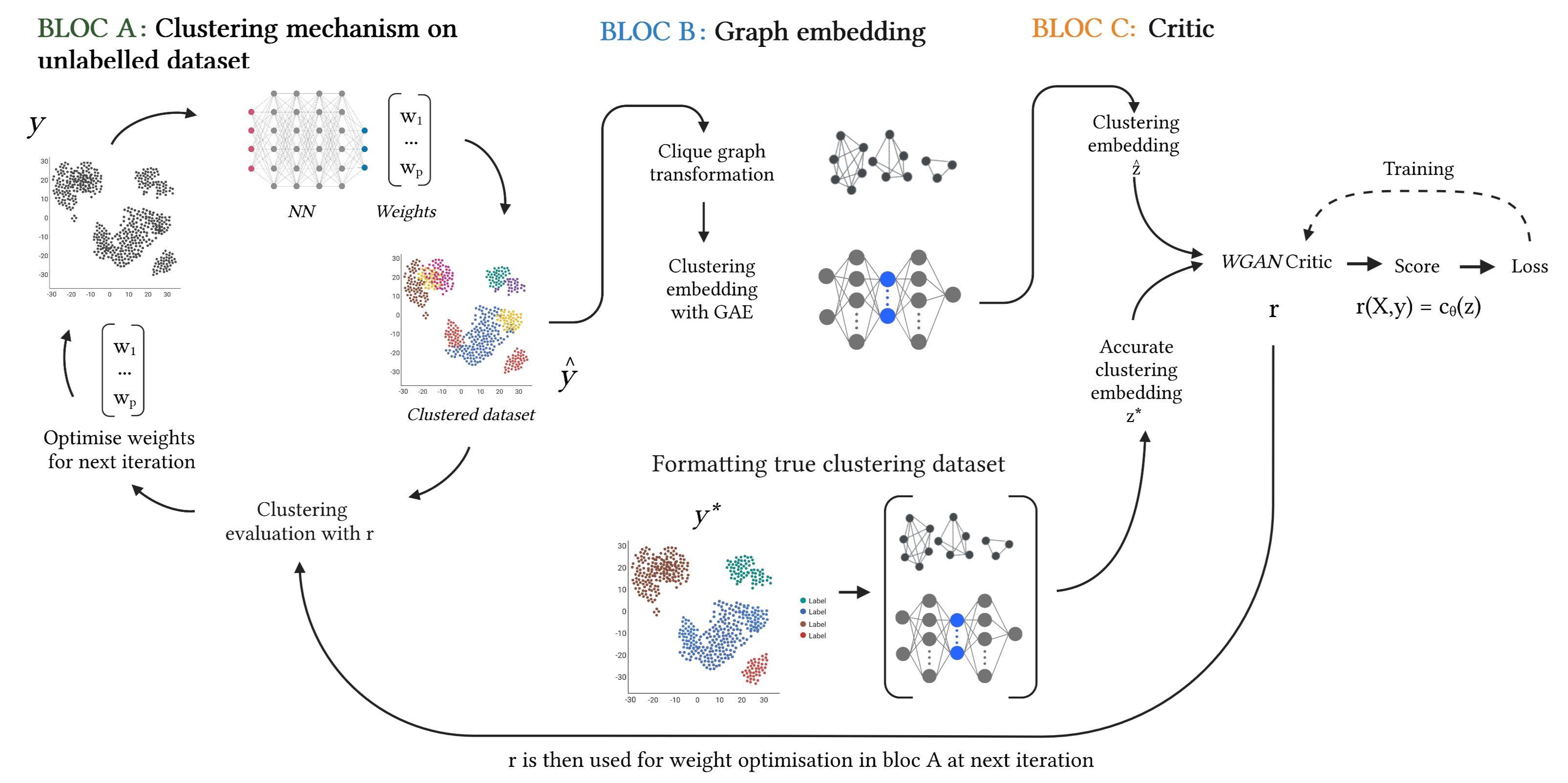}
	\caption{Our framework's 3 components: the clustering mechanism (A), the GAE (B) and the WGAN (C). (A) takes an unlabelled dataset $\mathbf {X}$ as input and outputs a clustering $\mathbf{\hat{y}}$ that maximises a metric $r$. $\mathbf{\hat{y}}$ is then turned into a graph $\mathcal{G}(\mathbf{X},\mathbf{\hat{y}})$ then into an embedding vector $\mathbf{\hat{z}}$ using (B). Same goes for the correctly labelled dataset, which is embedded as $\mathbf{\hat{z}^*}$. Then, (C), which is the metric itself, evaluates $\mathbf{\hat{z}}$ and $\mathbf{z}^*$ using $c_\theta$ and is trained to produce a new metric $r$ which is then used for (A) in the next iteration.}
    \label{framework}
\end{figure}

Therefore, the metric $r$ is actually the composition of two functions $g$ and $c_\theta$ (the scoring function from $\mathbb R^{\bf e}$ to $\mathbb R$). Our training procedure is structured around 3 blocs A, B and C detailed in next sections and depicted in figure \ref{framework} and is summarised in the following main steps:
\vspace{15mm}
\begin{enumerate}[{Bloc A}. step 1]
    \item Select a labelled dataset $(\bf{X},\bf{y}^*) \sim\mathcal{D}$ 
    \vspace{-2mm}
    \item Given a metric function $r$ (output from bloc B step 2, or initialised randomly), we perform a clustering of dataset $\bf X$: $\mathbf{\hat y} =\argmax_\mathbf{y} r(\mathbf{X},\mathbf{y})$ 
\end{enumerate}
\vspace{-2mm}
\begin{enumerate}[{Bloc B}. step 1]
    \item $\bf y^*$ and $\bf{\hat{y}}$ are represented as graphs where each clique represents a cluster.\vspace{-2mm}
    \item Graph convolutional autoencoders perform feature extraction from $\bf \hat{y}$ and $\bf y^*$ and output embeddings $\bf \hat{z}$ and $\bf z^*$ 
\end{enumerate}
\vspace{-2mm}
\begin{enumerate}[{Bloc C}. step 1]
    \item The metric $r$ is modelled by a WGAN critic that outputs evaluations of the clusterings: $r(\bf X,\bf y^*) = c_\theta(\bf z^*)$ and $r(\bf X,\bf \hat{y}) = c_\theta(\bf \hat z)$\vspace{-2mm}
    \item Train the model using the error between $r(\bf X,\bf y^*)$ and $r(\bf X,\bf \hat{y})$.
\end{enumerate}

\vspace{-3mm}

\begin{comment}
\begin{table}[!ht]
\centering
\caption{Summary of notations}
    \begin{tabular}{lp{6.3cm}}
    %\caption{Summary of notations}
		 %\hline
		 %Symbol
		 \hline
 $\mathbf X$ &  A dataset of $n$ points,  $x_i \in \mathbb{R}^{\bf d}$\\ 
$\mathbf y^*$ & True clustering (cluster labels) of $\mathbf X$, $\in \{1,\dots,k\}^n$ \\
$\mathbf y$ & A possible clustering of $\mathbf X$ \\
$\hat{\bf y}$ & Clustering retained after optimisation for a fixed function $r$\\
$\mathcal{M}_{n,m}$ & Set of matrices with $n$ lines and $m$ columns\\
 $\mathcal{G}(\mathbf X,\mathbf y)$ & Graph representing a clustered version of $\mathbf X$\\
 $A$ & An adjacency matrix \\
 $X$ & Feature matrix of $\mathbf X$. $X \in \mathcal{M}_{n,d} $\\
 $\bf z^*,\bf\hat{z}$ & Embedding, $\in \mathbb{R}^{\bf e}$, of $(\bf{X},\bf{y}^*)$, and $(\bf{X},\bf{y})$, respectively \\
 $r$ & Metric $\mathbb{R}^{\bf n\times d}\times \mathbb N^{\mathbf n} \mapsto \mathbb{R}$, scoring of the clustering \\
 CEM & Cross-entropy method\\
$\mathcal{S}$ & Set of all intermediate clustering solutions found through CEM \\[1ex] 
 \hline
\end{tabular}
\end{table}
\end{comment}

\vspace{-8mm}
\subsection{Clustering mechanism}\label{clustering}

We seek the most suitable optimisation algorithm for clustering given $r$. Considering a neural network that performs the clustering, we need to find its weights $w$ such that the metric is maximised (see equation \eqref{w}). The type of algorithm to use depends on the nature of the metric $r$ to optimise on. 

\begin{equation}\label{w}
   \text{CEM}_r(\mathbf X)\xrightarrow{\text{finds}} w^* = \argmax_w r(\mathbf{X},\mathbf{y}^w)
\end{equation}

Where $\mathbf y^w$ is a clustering obtained with the weights $w$. The metric is assumed to hold certain properties, discussed in \ref{critic}:

\begin{itemize}
    \item \textbf{Unique Maximum:} A unique optimal clustering.  $r$ has a unique maximum. 
    \vspace{-6mm}
    \item \textbf{Continuity\footnote{As a reminder, Let $T$ and $U$ be two topological spaces. A function $f:T\mapsto U$ is continuous in the open set definition if for every $t\in T$ and every open set $u$ containing $f(t)$, there exists a neighbourhood $v$ of $t$ such that $f(v)\subset u$.}}: Any two clusterings $\mathbf y$ and $\mathbf y'$ should be similar if $r(\mathbf y)$ and $r(\mathbf y')$ are close in $\mathbb{R}$ space. Hence, $r$ has to satisfy a continuity constraint.
\end{itemize}

%why is continuity so important
% ES take advantage of continuity to solve the problem while satisfying the "non differentiable" constraint

There is no guarantee that the best metric for the clustering task is differentiable. 
Given the above assumptions, conditions are favourable for evolutionary strategies (ES) to iteratively converge towards the optimal solution. Indeed, if $r$ is continuous and the series $((\mathbf{X},\mathbf{y}_1),\dots,(\mathbf{X},\mathbf{y}_p))$ converges towards $(\mathbf{X},\mathbf{y}^*)$ then $(r(\mathbf{X},\mathbf{y}_1),\dots,r(\mathbf{X},\mathbf{y}_p))$ converges towards $r(\mathbf{X},\mathbf{y}^*)$.
% CEM formulas and algorithm
We choose the Cross-Entropy Method (CEM) \cite{CEM}, a popular ES algorithm for its simplicity, to optimise the clustering neural network weights by solving Eq.\eqref{w} (algorithm \ref{Alg:CEM_algo}).

\begin{algorithm}[tb]
    \caption{CEM Algorithm}
    \label{Alg:CEM_algo}
\begin{algorithmic}
    \STATE \textbf{Input:} Dataset $X\in\mathbb R^{\bf{n} \times \bf{d}}$; score function $r$; $\mu \in \mathbb{R}^{\bf{d}}$ and $\sigma \in \mathbb{R}^{\bf{d}}$; elite percentage to retain $p$; $n$ samples of $w_i \sim \mathcal{N}(\mu,\text{diag}(\sigma))$; $T$ number of iterations
    \FOR{$\textnormal{iteration}=1$ {\bfseries to} $T$}
    \STATE Produce $n$ samples of neural network weights $w_i \sim \mathcal{N(\mu,\text{diag}(\sigma))}$
    \STATE Produce clusterings $y_i$ of $X$ using each $w_i$
    \STATE Evaluate $r_i = r(X,y_i)$
    \STATE Constitute the elite set of $p\%$ best $w_i$
    \STATE Fit a Gaussian distribution with diagonal covariance to the elite set and get a new $\mu_t$ and $\sigma_t$
    \ENDFOR
    \STATE {\bfseries return:} $\mu$, $w^*$
\end{algorithmic}
\end{algorithm}

\subsection{Graph based dataset embedding}

To capture global properties and be transferable across different datasets, we argue that it is necessary to input all the points of a dataset at once. Hence, instead of pairwise similarities between random pairs of points, we propose to get a representation of the relation between a bunch of neighbouring points. Thus, we represent each dataset by a graph structure $\mathcal{G}(\mathbf{X},\mathbf y)$ where each node corresponds to a point in $\mathbf{X}$ and where cliques represent clusters as shown in figure \ref{framework}. This representation takes the form of a feature matrix $X$ and an adjacency matrix $A$. 
Using $X$, and $A$, we embed the whole dataset into a vector $\bf z \in \mathbb{R}^{\mathbf e}$. To do so, we use graph autoencoders (GAE). Our implementation is based on \cite{GAE}.

\begin{comment}
Specifically, we have $\{X,A\} \mapsto \bf z$, under the following mechanism:

\begin{equation}
    \begin{aligned}
    GCN(X,A)= Relu(\Tilde{A}XW_0) = \Bar{X}\\
    \end{aligned}
\end{equation}

With $\Tilde{A}$ the symmetrically normalized adjacency matrix and $(W_0,W_1)$ the GCN weight matrices.

\begin{equation}
    \begin{aligned}
    z=\Tilde{A}\Bar{X}W_1
    \end{aligned}
\end{equation}

Finally, the decoder outputs a new adjacency matrix using the sigmoid function $\sigma$: 

\begin{equation}
    \begin{aligned}
    \hat{A}=\sigma(zz^T)
    \end{aligned}
\end{equation}

%The obtained embedding has to be tweaked a little to be independent of the number of points in the dataset
\end{comment}

We obtain $z \in \mathcal{M}_{n,m}$ which is dependent of the shape of the dataset (where $m$ is a user specified hyper-parameter). In order to make it independent from the number of points in $\mathcal{X}$, we turn the matrix $z$ into a square symmetrical one $z \xleftarrow{} z^Tz \in \mathcal{M}_{m,m}$. The final embedding corresponds to a flattened version of the principal triangular bloc of $z^Tz$, which shape is $\mathbf e=(\frac{m+1}{2},1)$. However, the scale of the output still depends on the number of points in the dataset. This could cause an issue when transferring to datasets with a vastly different number of data points. It should therefore require some regularisation; in order to simplify, we decided to use datasets with approximately the same number of points.

\subsection{A critic as a metric}\label{critic}

With embedded vectors of the same shape, we compare the clusterings proposed $\mathbf{\hat{z}}$ and the ground truth ones $\bf z$ using the metric $r$. $r$ is a function mapping an embedding vector $\mathbf z\in \mathbb R^{\mathbf e}$ to $\mathbb{R}$, we therefore parameterise it as:
\begin{equation}\label{large_state_reward}
r_\alpha(\mathbf X,\mathbf y)=r_\alpha(\mathbf z)=\alpha_1\phi_1(\mathbf z)+\alpha_2\phi_2(\mathbf z)+...+\alpha_h\phi_h(\mathbf z)
\end{equation}

Where $\phi_j(\mathbf z)\in \mathbb R$. As per \cite{Russell}, learning a viable metric is possible provided both the following constraints: (1) maximising the difference between the quality of the optimal decision and the quality of the second best; (2) minimising the amplitude of the metric function as using small values encourages the metric function to be simpler, similar to regularisation in supervised learning.

When maximising the metric difference between the two clusterings that have the highest scores, we get a similarity score as in traditional metric learning problems. The problem is formulated by equation \eqref{general_optimization} where  $\mathcal{S}$ is a set of solutions (i.e., clustering proposals) found using $r_\alpha$ and $\mathbf{y}^*$ is the true clustering, $\mathbf{y}^{\text{max}}$ is the best solution found in $\mathcal{S}$: $\mathbf{y}^{\text{max}} = \argmax_{\mathbf{y}\in\mathcal S}r_\alpha(\mathbf X, \mathbf{y})$. 
\begin{equation}\label{general_optimization}
    \begin{aligned}
    \min_\alpha r_\alpha(\mathbf X, \mathbf y^*) & -\max_\alpha \min_{\mathbf y'\in \mathcal S\setminus \mathbf y^{\text{max}}} r_\alpha(\mathbf X,\mathbf y^{\text{max}})-r_\alpha(\mathbf X,\mathbf y')\\
    & \quad \text{s.t} \quad \mathbf{y}^*=\argmax_{\mathbf{y}\in \mathbf{Y}}r(\mathbf{y})
    \end{aligned}
\end{equation}
%\vspace{-6mm}
\begin{algorithm}[h!]
\footnotesize
\caption{Critic2Metric (C2M)}\label{Complete_algo}
\SetAlgoLined
\KwInput{$b$: batch size, $epoch$: number of epochs; $p$: percentage of elite weights to keep; $iteration$: number of CEM iterations; $population$: number of weights to generate; $\mu \in \mathbb{R}^d$: CEM mean; $\sigma \in \mathbb{R}^d$: CEM standard deviation, $\theta$: critic's weights}
 \For{$n=1$ {\bfseries to} epoch}{
   \For{$k=1$ {\bfseries to} b}{
    Sample $(\mathbf X_{k},\mathbf y_k^*) \sim \mathcal D $ a correctly labelled dataset\\
    Generate ground truth  embeddings  $\mathbf z_{(\mathbf X_{k},\mathbf y_k^*)}=GAE(\mathcal{G}(\mathbf X_k,\mathbf y_k^*))$ \\
%    Create empty list of weights, $L = \{\}$ \;
    Initialise clustering neural network weights $\{w_j\}_{j=1}^{population}$  \\
%    \For{$j=1$ {\bfseries to} population}{
%     Initialize clustering neural network weights $\theta_j$ \;
%     Append $\theta_j$ to $L$}
     \For{$i=1$ {\bfseries to} iteration}{
     \For{$j=1$ {\bfseries to} population}
     {Generate clusterings $\mathbf{\hat{y}}_k^{w_j}$ \\
     Convert $\mathbf{\hat{y}}_k^{w_j}$ into a graph\\
     $\mathbf z_{(\mathbf X_{k},\mathbf {\hat{y}}_k^{w_j})}=  GAE(\mathcal{G}(\mathbf X_k,\hat{\mathbf y}_k^{w_j}))$ \\
     Evaluate: $r(\mathbf X_k,\hat{\mathbf y}_k^{w_j}) = c_\theta(\mathbf z_{(\mathbf X_{k},\mathbf {\hat{y}}_k^{w_j})})$}
     Keep proportion $p$ of best weights $w_p$ \\
     $w^* \xleftarrow{} \text{CEM}(w_p, \mu, \sigma)$}
     Generate clustering $\mathbf{y}_k^{w^*}$\\
     $\mathbf z_{(\mathbf X_{k},\mathbf {\hat{y}}_k^{w^*})} = GAE(\mathcal{G}(\mathbf X_k,\hat{\mathbf y}_k^{w^*}))$ \\
     Train critic as in \cite{WGAN} using $\mathbf z_{(\mathbf X_{k},\mathbf {\hat{y}}_k^{w^*})}$ and $\mathbf z_{(\mathbf X_{k},\mathbf y_k^*)}$ \;
    }}
\end{algorithm}
\vspace{-5mm}

To solve equation \eqref{general_optimization}, we %build on the approach of \cite{guided}, where the authors show that certain IRL methods are mathematically equivalent to GAN. In our 
use a GAN approach where the clustering mechanism (i.e., CEM) plays the role of the generator while a critic (i.e., metric learning model) plays the role of the discriminator. In a classic GAN, the discriminator only has to discriminate between real and false samples, making it use a cross entropy loss. With this kind of loss, and in our case, the discriminator quickly becomes too strong. Indeed, the score output by the discriminator becomes quickly polarised around 0 and 1.
\vspace{-1mm}

For this reason, we represent $r$ as the critic of a WGAN \cite{WGAN}. This critic scores the realness or fakeness of a given sample while respecting a smoothing constraint. The critic measures the distance between data distribution of the training dataset and the distribution observed in the generated samples. Since WGAN assumes that the optimal clustering provided is unique, the metric solution found by the critic satisfies equation \eqref{general_optimization} constraints. $r$ reaching a unique maximum while being continuous, the assumptions made in section \ref{clustering} are correctly addressed. 
To train the WGAN, we use the loss $\mathcal{L}$ in equation \eqref{WGAN_loss} where $\bf \hat{z}$ is the embedding vector of a proposed clustering and $\bf z$ is the embedding vector of the desired clustering. Our framework is detailed in algorithm \ref{Complete_algo}.
\vspace{-1mm}
\begin{equation}\label{WGAN_loss}
\mathcal{L}(\mathbf z^*,\mathbf {\hat{z}})=\max_{\theta}\mathbb{E}_{\mathbf z^*\sim p}[f_\theta(\mathbf z^*)] - \mathbb{E}_{\mathbf {\hat{z}}\sim p(\mathbf {\hat{z})}}[f_\theta(\mathbf {\hat{z}})]
\end{equation}

\section{Experiments}\label{sec:experiments}

%In this section, we carry out empirical evaluation of the proposed model using synthetic and real datasets. The purpose of this analysis is (1) to study the capacity of the model to provide a close to expected clustering under a learned metric, (2) to test its unsupervised clustering performance when transferring the metric to new different target datasets both in terms of domain and task.

\vspace*{-\baselineskip}
\begin{table*}[h!]
    \centering
 \begin{adjustbox}{width=\columnwidth, center}
 \begin{tabular}{||c || c || c || c || c || c || c || c || c || c ||} 
 \hline
 %Col1 & Col2 \\ [0.5ex] 
 \multicolumn{1}{||c|}{\textbf{Dataset family}} &
 \multicolumn{4}{||c|}{Synthetic data} &
 \multicolumn{3}{||c|}{MNIST} &
 \multicolumn{1}{||c|}{\begin{tabular}{@{}c@{}}Street view\\ house numbers\end{tabular}} &
 \multicolumn{1}{c||}{Omniglot} \\
 \hline
 \multicolumn{1}{||c|}{\textbf{Dataset}} &
 \multicolumn{1}{||c|}{Blob} &
 \multicolumn{1}{||c|}{Moon} &
 \multicolumn{1}{||c|}{Circles} &
 \multicolumn{1}{||c|}{\begin{tabular}{@{}c@{}}Aniso-\\ tropic\end{tabular} } &
 \multicolumn{1}{||c|}{\begin{tabular}{@{}c@{}}MNIST-digits\\ \cite{MNIST_digits}\end{tabular}} &
 \multicolumn{1}{||c|}{\begin{tabular}{@{}c@{}}letters MNIST\\ \cite{MNIST_letters}\end{tabular} } &
 \multicolumn{1}{||c|}{\begin{tabular}{@{}c@{}}fashion MNIST\\ \cite{fashion_MNIST}\end{tabular}} &
 \multicolumn{1}{||c|}{\begin{tabular}{@{}c@{}}SVHN\\  \cite{SVHN}\end{tabular}} &
 \multicolumn{1}{c||}{\begin{tabular}{@{}c@{}}Omniglot\\  \cite{omniglot}\end{tabular} } \\
 \hline
 \multicolumn{1}{||c|}{\textbf{Snapshot}} &
 \multicolumn{1}{||c|}{\raisebox{-\totalheight}{\includegraphics[width=20mm, height=20mm]{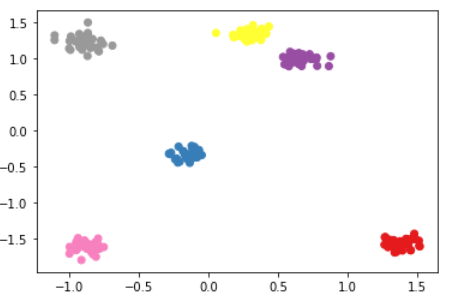}}} &
 \multicolumn{1}{||c|}{\raisebox{-\totalheight}{\includegraphics[width=20mm, height=20mm]{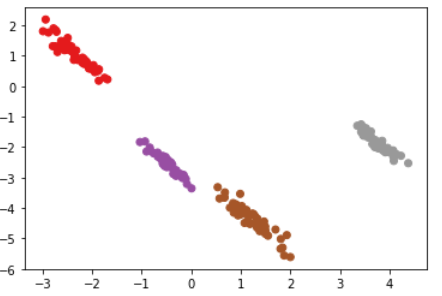}}} &
 \multicolumn{1}{||c|}{\raisebox{-\totalheight}{\includegraphics[width=20mm, height=20mm]{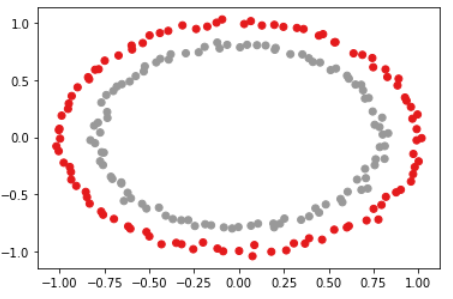}}} &
 \multicolumn{1}{||c|}{\raisebox{-\totalheight}{\includegraphics[width=20mm, height=20mm]{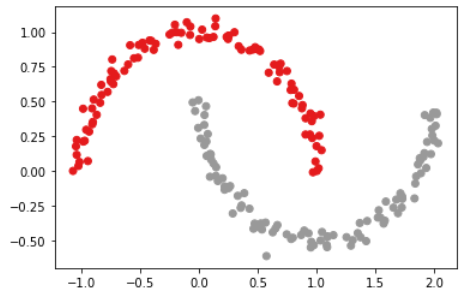}}} &
 \multicolumn{1}{||c|}{\raisebox{-\totalheight}{\includegraphics[width=20mm, height=20mm]{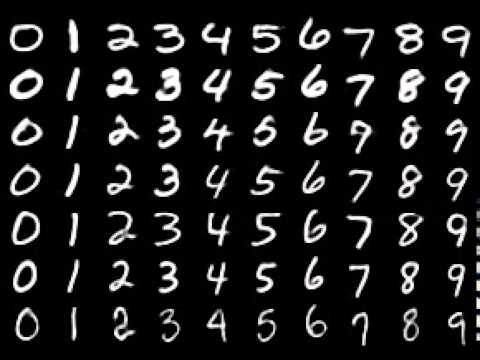}}} &
 \multicolumn{1}{||c|}{\raisebox{-\totalheight}{\includegraphics[width=20mm, height=20mm]{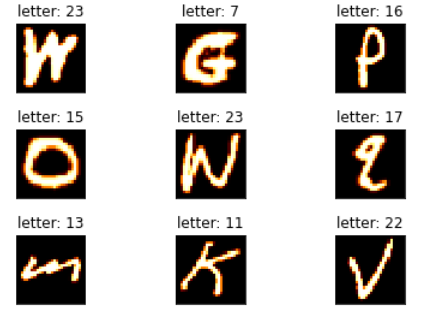}}} &
 \multicolumn{1}{||c|}{\raisebox{-\totalheight}{\includegraphics[width=20mm, height=20mm]{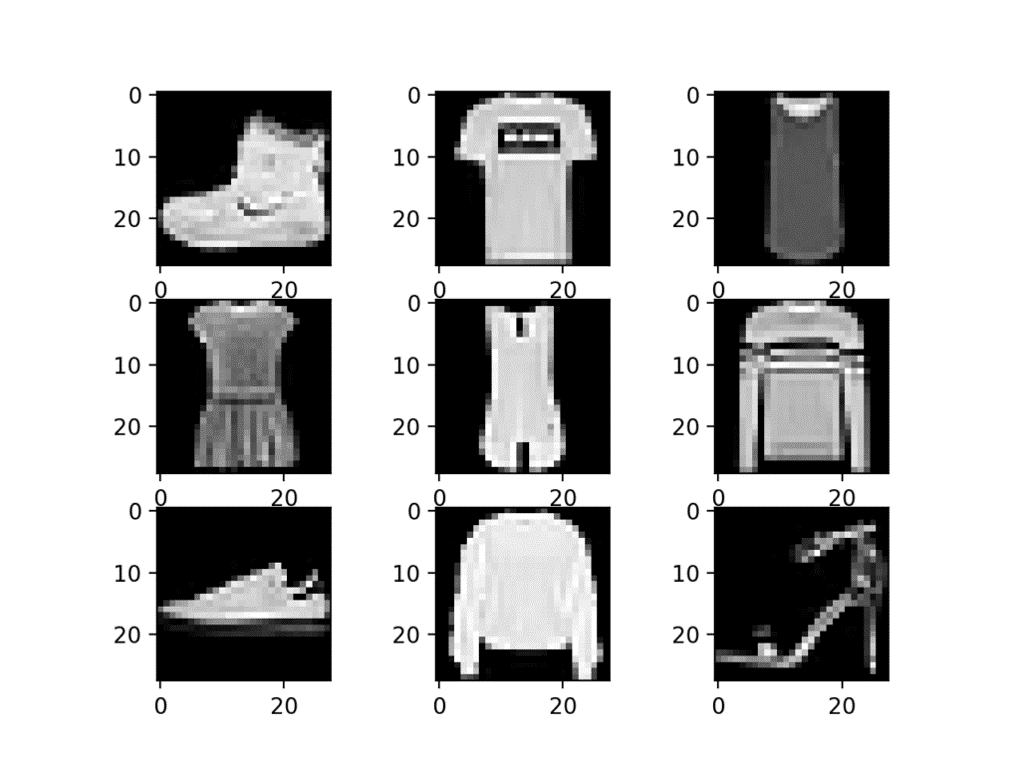}}} &
 \multicolumn{1}{||c|}{\raisebox{-\totalheight}{\includegraphics[width=20mm, height=20mm]{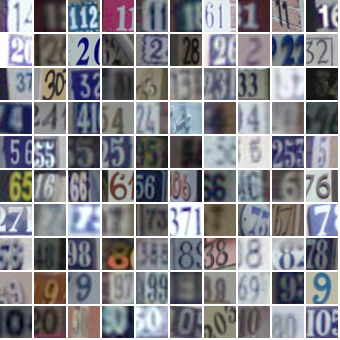}}} &
 \multicolumn{1}{c||}{\raisebox{-\totalheight}{\includegraphics[width=20mm, height=20mm]{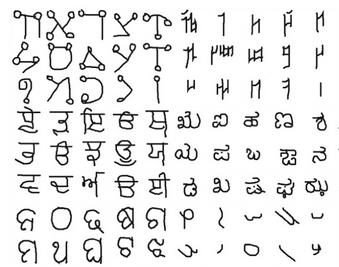}}} \\
 \hline
 \multicolumn{1}{||c|}{\textbf{\begin{tabular}{@{}c@{}}Feature\\ dimension\end{tabular} }} &
 \multicolumn{1}{||c|}{2} &
 \multicolumn{1}{||c|}{2} &
 \multicolumn{1}{||c|}{2} &
 \multicolumn{1}{||c|}{2} &
 \multicolumn{1}{||c|}{$28\times 28$} &
 \multicolumn{1}{||c|}{$28\times 28$} &
 \multicolumn{1}{||c|}{$28\times 28$} &
 \multicolumn{1}{||c|}{$32 \times 32$} &
 \multicolumn{1}{c||}{$105 \times 105$} \\
 \hline
 \multicolumn{1}{||c|}{\textbf{\begin{tabular}{@{}c@{}}Maximum number\\ of clusters\end{tabular}}} &
 \multicolumn{1}{||c|}{\begin{tabular}{@{}c@{}}9\\ (custom)\end{tabular}} &
 \multicolumn{1}{||c|}{\begin{tabular}{@{}c@{}}9\\ (custom)\end{tabular}} &
 \multicolumn{1}{||c|}{\begin{tabular}{@{}c@{}}9\\ (custom)\end{tabular}} &
 \multicolumn{1}{||c|}{\begin{tabular}{@{}c@{}}9\\ (custom)\end{tabular}} &
 \multicolumn{1}{||c|}{10} &
 \multicolumn{1}{||c|}{26} &
 \multicolumn{1}{||c|}{10} &
 \multicolumn{1}{||c|}{10} &
 \multicolumn{1}{c||}{47} \\
 \hline
 \multicolumn{1}{||c|}{\textbf{Size}} &
 \multicolumn{1}{||c|}{\begin{tabular}{@{}c@{}}200\\ (custom)\end{tabular}} &
 \multicolumn{1}{||c|}{\begin{tabular}{@{}c@{}}200\\ (custom)\end{tabular}} &
 \multicolumn{1}{||c|}{\begin{tabular}{@{}c@{}}200\\ (custom)\end{tabular}} &
 \multicolumn{1}{||c|}{\begin{tabular}{@{}c@{}}200\\ (custom)\end{tabular}} &
 \multicolumn{1}{||c|}{60000} &
 \multicolumn{1}{||c|}{145600} &
 \multicolumn{1}{||c|}{60000} &
 \multicolumn{1}{||c|}{73257} &
 \multicolumn{1}{c||}{32460} \\
 \hline
\end{tabular}
\end{adjustbox}
\caption{Datasets description}
\vspace{-8mm}
\label{tab:dataset}
\end{table*}
%\vspace{-2mm}

For empirical evaluation, we parameterise our framework as follows: The critic (block C in Fig~\ref{framework}) is a 5 layer network of sizes 256, 256, 512, 512, and 1 (output) neurons. All activation functions are LeakyRelu ($\alpha=0.2$) except last layer (no activation). RMSprop optimizer with $0.01$ initial learning rate and a decay rate of $0.95$. The CEM-trained neural network (bloc A in Fig~\ref{framework}) has 1 hidden layer of size 16 with Relu activation, and a final layer of size $k=50$ (the maximum number of clusters). The GAE (bloc B in Fig~\ref{framework}) has 2 hidden layers; sized 32 and 16 for synthetic datasets, and 100 and 50 for real datasets.

We choose datasets based on 3 main criteria: having a similar compatible format; datasets should be large enough to allow diversity in subsampling configurations to guarantee against overfitting; datasets should be similar to the ones used in our identified baseline literature. All used datasets are found in table \ref{tab:dataset}. 

For training, we construct $n$ sample datasets and their ground truth clustering, each containing 200 points drawn randomly from a set of 1500 points belonging to the training dataset. Each one of these datasets, along with their clustering is an input to our model. To test the learned metric, we construct 50 new sample datasets from datasets that are different from the training one (e.g., if we train the model on MNIST numbers, we will use datasets from MNIST letters or fashion to test the metric). The test sample datasets contain 200 points each for synthetic datasets and 1000 points each otherwise. The accuracies are then averaged accross the 50 test sample datasets.
To test the ability of the model to learn using only a few samples, we train it using 5 (few shots) and 20 datasets (standard), each containing a random number of clusters. For few shots trainings, we train the critic for 1 epoch and 10 epochs for standard trainings. 

To evaluate the clustering, we use Normalised-Mutual Information (NMI) \cite{NMI} and clustering accuracy (ACC) \cite{ACC}. NMI provides a normalised measure that is invariant to label permutations while ACC measures the one-to-one matching of labels. For clustering, we only need that the samples belonging to the same cluster are attributed the same label, independently from the label itself. However, since we want to analyse the behaviour of the metric learned through our framework, we are interested in seeing whether it is permutation invariant or not. Hence, we need the two measures.

\subsection{Results on 2D synthetic datasets}

Analysis on synthetic datasets (see table \ref{tab:dataset}) proves that our model behaves as expected. We do not compare our results to any baseline since existing unsupervised methods are well studied on them. 
We train our model using exclusively samples from blobs datasets. We then test the learned metric on the 4 different types of synthetic datasets (blobs, anisotropic, moons and circles). Results are displayed in table \ref{sci-kit_results}. We observe that the model obtains the best score on blobs since it is trained using this dataset. We can also notice that our model achieves high scores for the other types of datasets not included in training.
%\vspace*{-\baselineskip}
\begin{table} [h!]
\centering
\begin{tabular}{LCCCC}
\toprule
\multicolumn{1}{l}{Types of datasets} &
\multicolumn{2}{c}{Standard training}    &
\multicolumn{2}{c}{Few shots training}    \\ 
\cmidrule(lr){2-3}
\cmidrule(lr){4-5}

&
\multicolumn{1}{c}{ACC} &
\multicolumn{1}{c}{NMI}     &
\multicolumn{1}{c}{ACC} &
\multicolumn{1}{c}{NMI}     \\
\midrule

\text{Blobs} & 98.4\% & 0.980 & 97.3\% & 0.965\\
\text{Anisotropic} & 97.9\% & 0.967 & 97.2\% & 0.945\\
\text{Circles} & 91.7\% & 0.902 & 92.7\% & 0.900\\
\text{Moons} & 92.1\% & 0.929 & 92.8\% & 0.938\\
\bottomrule
\end{tabular}
\caption{Average ACC and NMI on synthetic test datasets.}
\vspace{-5mm}
\label{sci-kit_results}
\end{table}
%\vspace{-6mm}

Our model succeeds in clustering datasets presenting non linear boundaries like circles while blobs datasets used in training are all linearly separable. Hence, the model learns intrinsic properties of training dataset that are not portrayed in the initial dataset structure, and thus that the metric appears to be transferable.

\begin{figure}[h!]
    \centering
    \includegraphics[width=0.6\columnwidth]{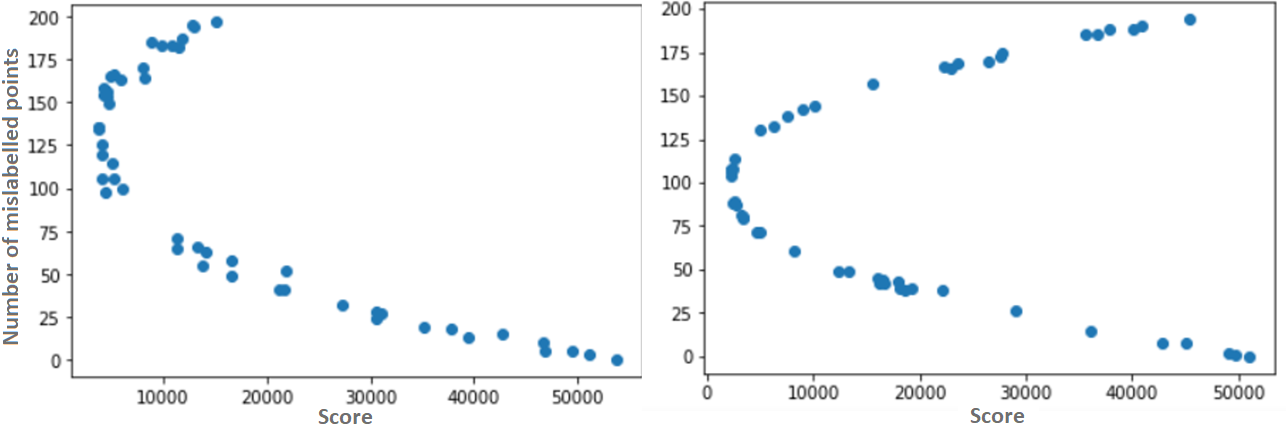}
	\caption{Metric values (i.e., scores given by the critic) for several clusterings of a dataset. Plots are from an anisotropic dataset (left) and a moons dataset (right). In a 2 cluster case (right), the formula used to compute mislabelled points has been made sensitive to label permutation to verify if permuted labels can fool the critic. The critic assigns a high score either when all the labels match the given ground truth or when all the labels are permuted (which again does not affect the correctness of the clustering)}
    \vspace{-2mm}
    \label{ablation}
\end{figure}

\textbf{Critic's ablation study}. To test if the critic behaves as expected, i.e., grades the clustering proposals proportionally to their quality, we test it on wrongly labelled datasets to see if the score decreases with the number of mislabelled points. We consider 50 datasets from each type of synthetic datasets, create 50 different copies and mislabel a random number of points in each copy. A typical result is displayed in figure \ref{ablation} and shows that the critic effectively outputs an ordering metric as the score increases when the number of mislabelled points decreases, reaching its maximum when there is no mislabelled point. This shows that the metric satisfies the constraints stated in equation \ref{general_optimization}. 
\vspace{-1mm}

\vspace{-6mm}
An interesting behaviour is shown in figure \ref{ablation}. Recall that since we are in the context of a clustering problem, we only need for the samples belonging to the same cluster to get the same label, independently from the cluster label itself. Thus, the formula used to compute mislabelled points has been made sensitive to label permutation to verify if permuted labels can fool the critic. For instance, in a 2 clusters case, one can switch the labels of all points in each cluster and still get the maximum score. Switching all labels makes all the points wrongly labelled compared to the given ground truth but nonetheless the clustering itself remains true. This explains the rounded shape in figure \ref{ablation} where the used datasets in the right panel only consisted of 2 clusters. The critic assigns a high score either when all the labels match the given ground truth or when all the labels are permuted (which does not affect the correctness of the clustering). 

\vspace{-3mm}
\subsection{Results on MNIST datasets}\label{MNIST_section}
\vspace{-1mm}
MNIST datasets give similar results both in terms of ACC and NMI on all test datasets regardless of the used training dataset (see table \ref{MNIST_result}). Hence, the model effectively capture implicit features that are dataset independent. While standard training shows better results, the few shots training has close performance.

%\vspace{-4mm}

\begin{table}[h!]
%\begin{adjustwidth}{-3cm}{-3cm}
%\begin{subtable}[t]{0.5\linewidth}
%\begin{table} [h!]
\centering
%\begin{adjustbox}{width=0.5\columnwidth}
\begin{tabular}{LCCCCCC}
\toprule
\multicolumn{1}{l}{Training Dataset} &
\multicolumn{6}{c}{Testing Dataset} \\
%\multicolumn{2}{c}{Few shots training}    \\ 
\cmidrule(lr){2-7} 
%\cmidrule(lr){4-5}
\multicolumn{1}{c}{} &
\multicolumn{2}{c}{Numbers} &
\multicolumn{2}{c}{Letters} &
\multicolumn{2}{c}{Fashion} \\
\cmidrule(lr){2-3} 
\cmidrule(lr){4-5} 
\cmidrule(lr){6-7} 
\multicolumn{1}{c}{} &
\multicolumn{1}{c}{ACC} &
\multicolumn{1}{c}{NMI} &
\multicolumn{1}{c}{ACC} &
\multicolumn{1}{c}{NMI} &
\multicolumn{1}{c}{ACC} &
\multicolumn{1}{c}{NMI} \\
\midrule
\text{Numbers (standard)} & 72.3\% & 0.733 & 81.3\% & 0.861 & 65.2\% & 0.792 \\
\text{Numbers (few shots)} & 68.5\% & 0.801 & 79.0\% & 0.821 & 61.8\% & 0.672 \\
\text{Letters (standard)} & 75.9\% & 0.772 & 83.7\% & 0.854 & 67.5\% & 0.800 \\
\text{Letters (few shots)} & 69.8\% & 0.812 & 78.7\% & 0.806 & 60.9\% & 0.641 \\
\text{Fashion  (standard)} & 70.6\% & 0.706 & 83.4\% & 0.858 & 72.5\% & 0.762 \\
\text{Fashion (few shots)} & 70.1\% & 0.690 & 82.1\% & 0.834 & 70.7\% & 0.697 \\
\bottomrule
\end{tabular}
 %\end{adjustbox}
\caption{Mean clustering performance on MNIST dataset.}
\label{MNIST_result}
\end{table}

\vspace{-12mm}

\begin{table}[h!]
\centering
\begin{tabular}{LCCCCCC}
\toprule
\multicolumn{1}{l}{Training Dataset} &
\multicolumn{6}{c}{Testing Dataset} \\
%\multicolumn{2}{c}{Few shots training}    \\ 
\cmidrule(lr){2-7} 
%\cmidrule(lr){4-5}
\multicolumn{1}{c}{} &
\multicolumn{2}{c}{Numbers} &
\multicolumn{2}{c}{Letters} &
\multicolumn{2}{c}{Fashion} \\
\cmidrule(lr){2-3}
\cmidrule(lr){4-5}
\cmidrule(lr){6-7}
\multicolumn{1}{c}{} &
\multicolumn{1}{c}{Best} &
\multicolumn{1}{c}{Top 3} &
\multicolumn{1}{c}{Best} &
\multicolumn{1}{c}{Top 3} &
\multicolumn{1}{c}{Best} &
\multicolumn{1}{c}{Top 3} \\
\midrule
\text{Numbers (standard)} & 78.3\% & 92.5\% & 86.0\% & 97.5\% & 69.2\% & 87.2\%\\
\text{Numbers (few shots)} & 75.8\% & 82.1\% & 83.3\% & 92.0\% & 65.1\% & 83.9\% \\
\text{Letters (standard)} & 77.4\% & 89.2\% & 88.8\% & 96.4\% & 70.2\% & 86.7\%\\
\text{Letters (few shots)} & 73.1\% & 80.6\% & 85.1\% & 91.5\% & 61.0\% & 76.3\% \\
\text{Fashion (standard} & 70.1\% & 83.1\% & 85.0\% & 98.6\% & 76.9\% & 94.7\%\\
\text{Fashion (few shots)} & 67.9\% & 77.4\% & 83.5\% & 95.3\% & 70.2\% & 88.0\%\\
\bottomrule
\end{tabular}
%\end{adjustbox}
\caption{Critic based performance assessment: Best corresponds to the percentage of times the critic gives the best score to the desired solution. Top 3 is when this solution is among the 3 highest scores.}
\label{MNIST_theoretic}
%\caption{Results on MNIST datasets}
\vspace{-4mm}
\end{table}

\vspace{-2mm}

Table \ref{MNIST_theoretic} shows the percentage of times the critic attributes the best score to the desired solution. It shows that ES algorithm choice has a significant impact on the overall performance. Even with a metric that attributes the best score to the desired clustering, the CEM may be stuck in a local optimum and fails to reconstruct back the desired clustering. Hence, a better optimisation can enhance the performance shown in table \ref{MNIST_result} closer to the one presented in table \ref{MNIST_theoretic}.

%This means that a better optimization tool is needed to push ACC and NMI shown in table \ref{MNIST_result} closer to the performance shown in table \ref{MNIST_theoretic}.

\subsection{Comparative study}

We compare our approach with baseline methods from the literature  
(table \ref{comparative_results}). For some methods, we followed the procedure in \cite{transfer_clustering} and used their backbone neural network as a pairwise similarity metric. Table \ref{Results_SVHN} reports results when training on SVHN and testing on MNIST numbers. We obtain close ACC values to CCN and ATDA \cite{ATDA}. These methods uses Omniglot as an auxiliary dataset to learn a pairwise similarity function, which is not required for our model. Our model only uses a small fraction of SVHN, has shallow networks and does not require any adaptation to its loss function to achieve comparable results. Finally, other cited methods require the number of clusters as an a priori indication. We achieve comparable results without needing this information. When the loss adaptation through Omniglot is discarded (denoted source-only in table \ref{Results_SVHN}), or if the number of clusters is not given, their accuracy falls and our model surpasses them by a margin.

\vspace{-6mm}

%\vspace*{-\baselineskip}
\begin{table}[h!]
\begin{subtable}[c]{0.5\textwidth}
\centering
\begin{tabular}{LCC}
    \toprule
    \text{Method} & \multicolumn{2}{c}{\text{ACC}} \\
    \midrule
    & \text{Loss Adaptation} & \text{Source Only}\\
     \midrule
    \text{DANN \cite{DANN}}   & 73.9\% & 54.9\%\\
    \text{LTR \cite{LTR}}    & 78.8\% & 54.9\%\\
    \text{ATDA \cite{ATDA}} & 86.2\% & 70.1\%\\
    \text{CCN \cite{transfer_clustering}} & 89.1\% & 52\%\\
    \text{Ours (standard)} & - & 84.3\% \\
    \text{Ours (few shots)} & - & 81.4\% \\
    \bottomrule
    \end{tabular}
\subcaption{Unsupervised cross-task transfer from SVHN to MNIST digits.}
\vspace{-2mm}
\label{Results_SVHN}
\end{subtable}
\begin{subtable}[c]{0.5\textwidth}
\centering
\begin{tabular}{LCC}
         \toprule
         \text{Method} & \text{ACC} & \text{NMI} \\
         \midrule
         \text{k-means} & 18.9\% & 0.464 \\
         \text{CSP \cite{CSP}} & 65.4\% & 0.812 \\
         \text{MPCK-means \cite{mpckmeans}} & 53.9\% & 0.816 \\
         \text{CCN \cite{transfer_clustering}} & 78.18\% & 0.874 \\
         \text{DTC \cite{learn}} & 87.0\% & 0.945 \\
          \text{Autonovel \cite{autonovel}} & 85.4\% & - \\
         \text{Ours (standard)} & 83.4\% & 0.891 \\
         \bottomrule
    \end{tabular}
\subcaption{Unsupervised cross-task transfer from $\text{Omniglot}_\text{train}$ to $\text{Omniglot}_\text{test}$ ($k=100$ for all).}
\vspace{-2mm}
\label{Omniglot_results}
\end{subtable}
\caption{Comparative clustering performance}
\vspace{-8mm}
\label{comparative_results}
\end{table}

Table \ref{Omniglot_results} reports results when training on $\text{Omniglot}_\text{train}$ and testing on $\text{Omniglot}_\text{test}$. Values are averaged across $20$ alphabets which have $20$ to $47$ letters. We set the maximum number of clusters $k=100$. When the number of clusters is unknown, we get an ACC score relatively close to DTC and Autonovel. Compared to these two approaches, our method bears several significant advantages:
\begin{itemize}
\vspace{-2mm}
    \item \textbf{Deep Networks}: DTC and Autonovel used Resnets as a backbone which are very deep networks while we only used shallow networks (2 layers maximum)
    \vspace{-6mm}
    \item \textbf{Pairwise similarity}: in Autonovel the authors used a pairwise similarity statistic between datasets instances which we aimed to avoid due to its significant computational bottleneck. Moreover, this metric is recalculated after each training epoch, which adds more complexity.
    \vspace{-2mm}
    \item \textbf{Vision tasks:} While DTC can only handle vision tasks, we present a more general framework which includes vision but also tabular datasets.
    \vspace{-2mm}
    \item \textbf{Number of classes}: DTC and Autonovel used the labelled dataset as a probe dataset, and estimates the number of classes iteratively, and when the labelled clusters are correctly recovered, they used the ACC metric to keep the best clustering. This approach is effective, but requires access to the labelled dataset at inference time to estimate the number of classes. This is a shortcoming (memory or privacy limitations). Our approach does not require the labelled dataset once the metric is learned. Our metric automatically estimates the number of clusters required to any new unlabelled dataset. 
\end{itemize}

\vspace{-2mm}
\section{Conclusion}\label{sec:discussion}
We presented a framework for cross domain/task clustering by learning a transferable metric. This framework consisted of ES methods, and GAE alongside a critic. Our model extracts dataset-independent features from labelled datasets that characterise a given clustering, performs the clustering and grades its quality. We showed successful results using only small datasets and relatively shallow architectures. Moreover, there is more room for improvement. Indeed, since our framework is composed of 3 different blocs (CEM, GAE, critic), overall efficiency can be enhanced by independently improving each bloc (i.e replacing CEM).

In future work, we will study the criteria that determine why some auxiliary datasets are more resourceful than others given a target dataset. In our case, this means to study for instance why using the MNIST letters dataset as training allowed a better performance on Fashion MNIST than when using MNIST numbers. This would allow to deliver a minimum performance guarantee at inference time by creating a transferability measure between datasets.

\textbf{Acknowledgements:} We gratefully acknowledge Orianne Debeaupuis for making the figure. We also acknowledge computing support from
NVIDIA. This work was supported by funds from the French Program "Investissements d'Avenir".

\vspace{-4mm}
\bibliographystyle{splncs04} 
\bibliography{Bibliography}
\end{document}